\documentclass[10pt,twocolumn,letterpaper]{article}

\usepackage{wacv}              

\definecolor{wacvblue}{rgb}{0.21,0.49,0.74}
\usepackage[pagebackref,breaklinks,colorlinks,allcolors=wacvblue]{hyperref}

\definecolor{myred}{RGB}{249, 221, 223} 
\definecolor{myblue}{RGB}{221, 222, 237}
\usepackage{multirow}
\usepackage{makecell}
\usepackage[table]{xcolor}

\def\wacvPaperID{2012} 
\def\confName{WACV}
\def\confYear{2026}

\title{Evaluating and Preserving High-level Fidelity in Super-Resolution}

\author{Josep M. Rocafort$^{1,2}$, Shaolin Su$^{1}$,
Alexandra Gomez-Villa$^{1,2}$,
Javier Vazquez-Corral$^{1,2}$
\\$^1$Computer Vision Center
\\$^2$Universitat Autonoma de Barcelona
}

\begin{document}

\twocolumn[{%
\renewcommand\twocolumn[1][]{#1}%
\maketitle
    \centering
    \includegraphics[width=1\linewidth]{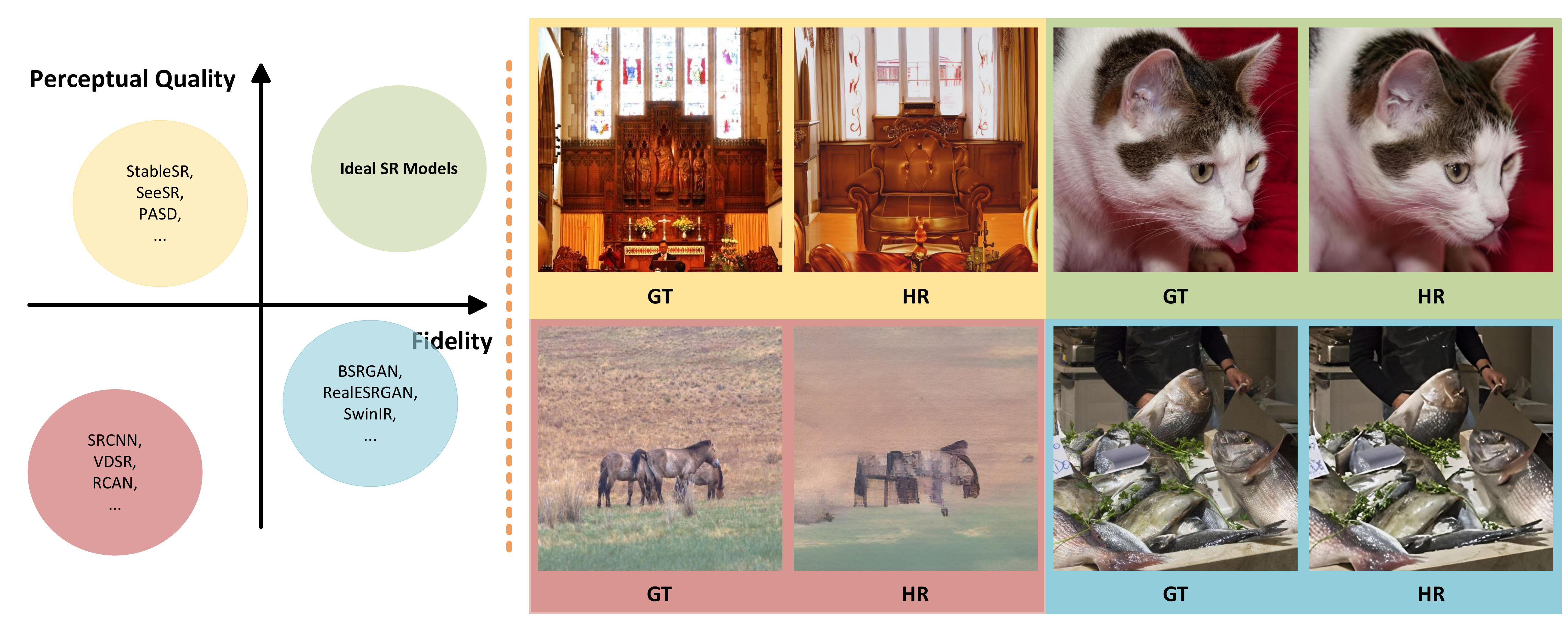}
    \captionof{figure}{We show that when evaluating SR models, high-level fidelity can reflect complementary aspects of images from the generally assessed image visual quality. Left: an illustration that, except for perceptual visual quality (evaluated vertically), models can perform differently in terms of preserving fidelity (evaluated horizontally). Right: SR examples corresponding to four quadrants, note that recent diffusion-based SR models can achieve high visual quality yet drastically change image fidelity (up-left), leading to less convincing results. Please zoom in for better view.}
    \label{fig:wacv_teaser}
    \vspace{1em}
}]

\begin{abstract}
  Recent image Super-Resolution (SR) models are achieving impressive effects in reconstructing details and delivering visually pleasant outputs. However, the overpowering generative ability can sometimes hallucinate and thus change the image content despite gaining high visual quality. This type of high-level change can be easily identified by humans yet not well-studied in existing low-level image quality metrics. In this paper, we establish the importance of measuring high-level fidelity for SR models as a complementary criterion to reveal the reliability of generative SR models. We construct the first annotated dataset with fidelity scores from different SR models, and evaluate how state-of-the-art (SOTA) SR models actually perform in preserving high-level fidelity. Based on the dataset, we then analyze how existing image quality metrics correlate with fidelity measurement, and further show that this high-level task can be better addressed by foundation models. Finally, by fine-tuning SR models based on our fidelity feedback, we show that both semantic fidelity and perceptual quality can be improved, demonstrating the potential value of our proposed criteria, both in model evaluation and optimization. We will release the dataset, code, and models upon acceptance.
\end{abstract}
    
\section{Introduction}
\label{sec:intro}


In the field of super-resolution (SR), we aim to convert low-quality, low-resolution (LR) images into clean, high-resolution (HR) versions. The output HR images are expected to not only possess rich, sharp details, but also maintain content fidelity consistent with the original Ground Truth (GT) images \cite{zhang2018rcan,wang2021realesrgan,Liang2021SwinIRIR,wang2024exploiting,sun2024pisasr}. In earlier SR research, due to the limited model capability, the HR outputs usually show both inferior quality and fidelity, \textit{e.g.} messy textures or vague structures. Thus, quality metrics, either perception metrics (\textit{e.g.} quality oriented metrics such as LPIPS \cite{zhang2018lpips} and DISTS \cite{ding2020iqa}) or distortion metrics (\textit{e.g.} fidelity oriented metrics such as PSNR and SSIM \cite{wang2004ssim}) are sufficient to measure SR model performances. However, with the advancement of generative models, especially recent diffusion models \cite{Rombach_2022_CVPR}, diffusion-based SR methods \cite{wu2024seesr,wang2024exploiting} become capable of producing high visual quality outputs that are perceptually comparable to or even better than GT images \cite{su2025rethinking,chen2025toward}. Although achieving high visual quality, the overpowering generative ability of these models can sometimes produce hallucinations which deviate from original image content, leading to even less convincing results than earlier SR models \cite{srgan, zhang2021designing, wang2021realesrgan}. As a result, current quality metrics become insufficient for measuring the fidelity aspects of recent advanced SR models, a crucial criteria which is however less noticed or studied in previous models and metrics.

In Figure \ref{fig:wacv_teaser}, we show how quality and fidelity reflect complementary aspects of SR models. As can be seen, though from visual quality aspect (vertically), most SR model performances can be distinguished, there are outputs demonstrating high visual quality while exhibiting content changes from the original (up-left), indicating model failures in producing convincing outputs. Therefore, to fully evaluate the representation capability of SR models, we highlight the importance of fidelity measurement, especially toward recent advancing diffusion-based SR methods. 
It is noteworthy that although some existing metrics such as PSNR and SSIM \cite{wang2004ssim} calculate pixel-wise distortions and are interpreted as ``fidelity'' metrics \cite{Blau_2018}, they focus on low-level aspects and lack discernment regarding high-level fidelity losses, an arising problem that emerges with recent diffusion-based SR models. 

\begin{figure}[t!]
    \centering
    \includegraphics[width=1\linewidth]{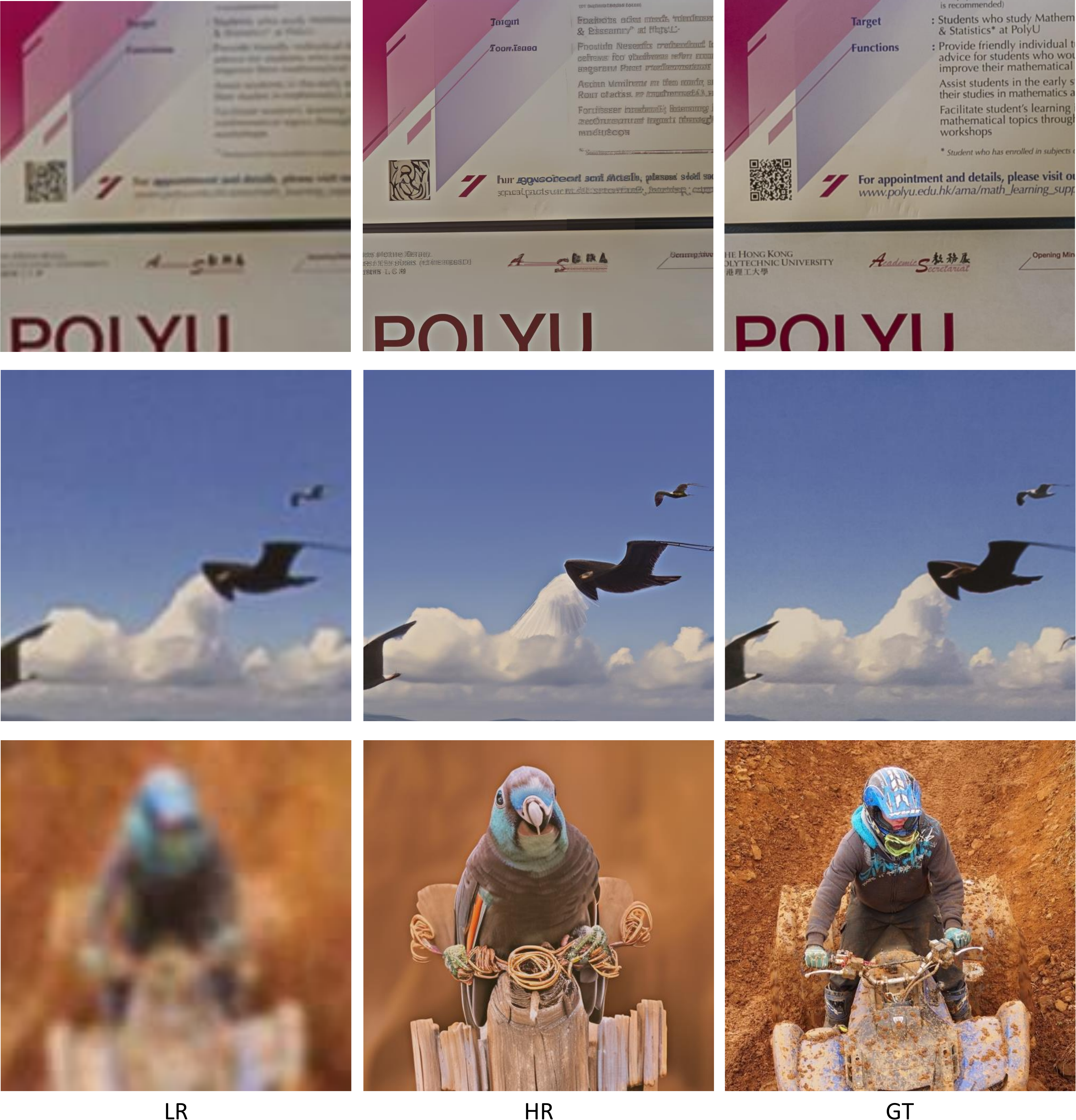}
    \caption{We show three different types of high-level fidelity changes. From top to bottom, fidelity change in details (texts and the QR code), fidelity changes in local structure (note the alteration of a cloud to the bird's wing reducing the reliability of the image), and fidelity changes in holistic semantics (the man in a quad is changed to a parakeet). Please zoom in for better view.}
    \label{fig:semantic_error_example}
\end{figure}

To better illustrate high-level fidelity, we define three types of fidelity losses that affect model representations, \textit{i.e.} fidelity losses in details, local structures, and change of holistic semantics. Fidelity losses in details denote small objects that encapsulate high-level information in a condensed form, such as texts, icons/symbols and QR code/barcode in images (Figure \ref{fig:semantic_error_example}, first row). Failures in reconstructing such information can sometimes result in the loss of key information in an image, \textit{e.g.} users are unable to scan a QR code from an HR image processed by a certain SR model. Fidelity losses in local structures include failing to reconstruct local objects such as human faces, shapes of creatures and man-made structures (buildings, cars and furniture which contain regular geometric shapes, Figure \ref{fig:semantic_error_example}, second row). The former two types of fidelity losses can exist only in a small region of images, thus being hard to capture by image quality or low-level fidelity measurements, yet representing a vital loss in high-level fidelity which can be easily identified by human observers. Fidelity losses in holistic semantics stand for the alteration of the whole object in the image (\textit{e.g.} changing a church to a sofa, changing a human to a parakeet, as shown in the third row of Figure \ref{fig:semantic_error_example}), due to excessively vague LR input or biased image prior learned by generation models. These alterations can generally produce high visual-quality results, however shift the information from the original images.

Based on the above motivation, in this paper, we study the high-level fidelity issue for SR models. In order to understand how different SR models are able to maintain high-level fidelity, we first select five state-of-the-art SR models, including one CNN + GAN model---BSRGAN \cite{zhang2021designing}, one Transformer + GAN model---SwinIR \cite{Liang2021SwinIRIR}, and three Diffusion based models---StableSR \cite{wang2024exploiting}, SeeSR \cite{wu2024seesr} and PASD \cite{yang2023pixelaware}, to test them on the classic $\times4$ SR problem. After a subjective study to annotate the high-level fidelity with GT images, we construct a dataset that contains HR outputs from different SR models and their fidelity evaluation scores. Based on the dataset, we are able to analyze both \textit{how SR models perform in maintaining image fidelity} and \textit{how image metrics perform in measuring model fidelity}. We show that different SR models can perform differently in terms of maintaining fidelity, and models achieving high visual quality may, in contrast, drastically change image information. By investigating image metrics, we show that widely adopted image quality assessment (IQA) metrics can only partly, or even poorly reflect fidelity changes. Furthermore, we show that foundation vision models outperform traditional IQA metrics in measuring fidelity changes, enabling their use as feedback signals to improve SR model fidelity through fine-tuning.

To summarize, the main contributions of this paper include:

\begin{itemize}
    \item We establish the importance of high-level fidelity evaluation as an essential complementary perspective to traditional low-level IQA metrics for assessing the reliability of generative SR models. 
    \item We construct the first annotated fidelity evaluation dataset for SR models, especially targeting diffusion-based ones. By analyzing 723 images produced by five different SR models, we evaluate how state-of-the-art SR models can perform in terms of maintaining high-level image fidelity.
    \item We show that foundation vision models outperform traditional IQA metrics in capturing high-level fidelity changes, and demonstrate that incorporating their feedback during SR fine-tuning simultaneously improves both semantic fidelity and perceptual quality.
\end{itemize}

\section{Related work}

We first review SOTA SR models and current image metrics that evaluate SR models. Since recently, there are also research studies that focus on evaluating the consistency of generative models, we review these works and illustrate the differences.

\subsection{Image Super-Resolution}

Early deep learning based SR methods adopt CNN to recover details. These approaches rely on end-to-end training and representative works include using plain CNN \cite{SRCNN}, residual connections \cite{Kim_2016_VDSR}, dense connections \cite{TongSRSkip} and attention mechanisms \cite{zhang2018rcan}. To enhance visual effects, following approaches \cite{srgan, zhang2021designing, wang2021realesrgan} employ GAN framework to model real-world image distribution, showing superior abilities in generating natural details. Benefited from the non-local characteristic, transformer-based SR models such as IPT \cite{chen2020IPT} and SwinIR \cite{Liang2021SwinIRIR} further improved model capabilities by referring to long-range dependencies. Recently, Stable-Diffusion models \cite{Rombach_2022_CVPR} in text-to-image tasks have shown their strong ability in learning large-scale image priors, many SR works such as StableSR \cite{wang2024exploiting}, SeeSR \cite{wu2024seesr}, PASD \cite{yang2023pixelaware},  TSD-SR \cite{dong2025tsd} and PiSA-SR \cite{sun2024pisasr} leverage its superior generative capability and achieve impressive results of reconstructed images. 

\subsection{Image Quality Metrics}

Current SR research employs visual quality metrics to evaluate their models. SSIM \cite{wang2004ssim} calculates local structural similarities with GT image, VIF \cite{vif} models image fidelity based on mutual information, LPIPS \cite{zhang2018lpips} and DISTS \cite{ding2020iqa} measure HR and GT image similarity in perceptual feature space of deep models. Due to the advancement of diffusion-based SR models, the evaluation of these models becomes more challenging, and \cite{su2025rethinking,chen2025toward} notice that there already exist cases in which the HR image quality can even exceed GTs, due to the limited quality of GTs collected in early years. Therefore, No-Reference (NR) IQA metrics such as NIQE\cite{mittal2012making} IL-NIQE \cite{zhang2015feature}, PI \cite{ma2017learning} and Clip-IQA \cite{WangCL23_clipiqa} are also applied to evaluate the perceptual quality of SR models. Nevertheless, when evaluating SR models, it remains unavoidable to use GTs to assess whether the HR results are reliable and convincing, especially for diffusion-based SR models which are prone to hallucinate images. Therefore, different with measuring quality, we propose to evaluate SR models from another complementary perspective, that is, the fidelity maintenance power of SR models.

\subsection{Metrics to Evaluate Generative Models}

For generative models, fidelity and diversity are two general criterion to evaluate. IS \cite{salimans2016improved} measures class diversity of the generated images. FID \cite{heusel2017gans} and KID \cite{binkowski2018demystifying} calculates the distance between the generated image and real-world image distribution as fidelity metrics. Note that the two fidelity metrics \cite{heusel2017gans,binkowski2018demystifying} focus on measuring the holistic similarity between distributions to assess the overall model capability, whereas in this paper, we emphasize more on fine-grained fidelity, \textit{i.e.} fidelity for individual samples and even in local image regions. Due to the success of Diffusion Models, recently proposed metrics also elevate to more stringent standards. Q-Eval-Score \cite{zhang2025q} measures both quality and alignment between the generated images and the texts prompts, \cite{lim_Evaluating} measures the factuality of generated
images through visual question answering. One recent work which is related with ours is JOINT \cite{Chen_Studyof}, where they measure the naturalness and rationality of the AI generated images. However, JOINT operates in an NR manner, while we target scenarios where GTs are available, thus focusing more on fidelity with GT instead of rationality without a reference (think of a case where the SR model changes the holistic semantic while the result still maintains high rationality). In Table \ref{tab:1}, we summarize existing image metrics and illustrate how the proposed task differs from them.

\begin{table}[t!]
\centering
\caption{Summary of existing image metrics and the proposed high-level fidelity task.}
\begin{tabular}{l|ll}
\hline
Method       & Task                & Scheme       \\ \hline
SSIM         & Visual quality      & FR           \\
LPIPS        & Visual quality      & FR           \\
Clip-IQA     & Visual quality      & NR           \\
FID          & Image fidelity      & Distribution \\
Q-Eval-Score & Text-image fidelity & TR           \\
I-HallA      & Text-image fidelity & TR           \\
JOINT        & Image naturalness      & NR           \\ \hline
Proposed     & Image fidelity      & FR           \\ \hline
\end{tabular}
\label{tab:1}
\end{table}

\section{High-Level Fidelity of SR Models}


Before delving into details, it is important to illustrate why we specifically focus on high-level fidelity. As is well known, SR is an ill-posed problem and there exists diverse solutions corresponding to one LR input \cite{VarSR}. This inherent characteristic of the SR problem seems to contradict our purpose of measuring the definite fidelity. However, in most cases, such diversity is only tolerable for scenarios where not much information is encoded \cite{VarSR,chan2021glean}, \textit{e.g.} grass in the background, texture of a sweater, hair strand of a person \textit{etc}. When evaluating the diverse results by referencing GTs, these alterations will be subtle to perceive, and human observers will rate them as plausible and good outputs. Therefore, we argue that in order to evaluate the convincement of SR models while being compatible with the inherent ill-posed characteristic, we allow fidelity changes on low-level and imperceptible solutions, and only focus on the fidelity errors that can be easily identified by human users, which are generally, image regions that encode high-level semantics.

After clarifying our objective, in the following, we construct an SR dataset with human-rated high-level fidelity annotations, and evaluate SOTA SR models based on the dataset.

\subsection{Dataset Construction}


As noted in \cite{su2025rethinking,chen2025toward}, GT images in current SR evaluations can contain distortions. To ensure our fidelity measurements are robust to such quality variations, we select images covering diverse quality levels for our dataset. Specifically, we use images from the KonIQ-10k dataset \cite{koniq10k} as ground truth references.
\begin{figure}
    \centering
    \includegraphics[width=1\linewidth]{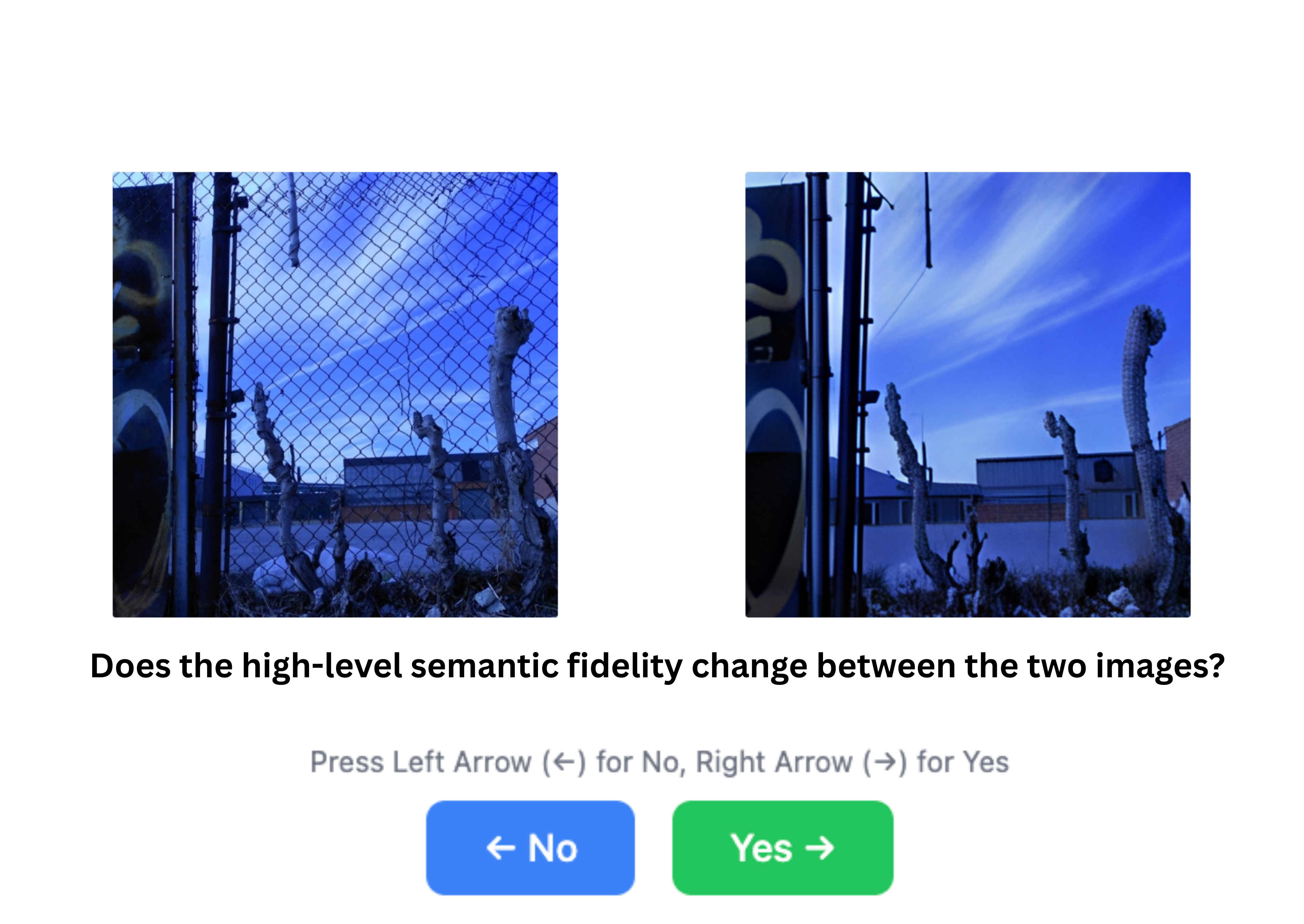}
    \caption{Our custom interface for the user study. Users are shown two images side to side. One of the images is the Ground Truth and the other is the output of a Super-Resolution model. Users are asked whether there is a high-level fidelity difference between both images, and are allowed to answer either ``Yes'' or ``No''.}
    \label{fig:placeholder}
\end{figure}


In order to construct a diverse dataset we downscale the images by 4 and applied different levels of degradation following the BSRGAN degradation pipeline \cite{zhang2021designing}. In the following, we employ five different SOTA SR models, including three diffusion based \cite{wu2024seesr,wang2024exploiting, yang2023pixelaware}, one ViT + GAN \cite{Liang2021SwinIRIR} and one CNN + GAN model  \cite{zhang2021designing}, and test them on our constructed LR images. All the models are implemented using their officially released codes and pre-trained weights. Finally, we selected 723 LR-GT image pairs based on SR model outputs and PE core\cite{bolya2025PerceptionEncoder} GT-SR cosine similarity scores to ensure an even distribution across different similarity levels. We select PE-core as given its contrastive pretraining task, we expect it to be better correlated with high-level semantic fidelity evaluation. 
Our dataset is comprised of 148 BSRGAN images, 148 StableSR images, 144 PASD images, 142 SeeSR images and 141 SwinIR images.



Based on these images, we then conduct a subjective user study to annotate the high-level semantic changes of various SR models. We develop an online platform which allows participants to log in and complete the experiment from their own devices. Users are shown one HR-GT image pair, placed side-by-side at a time and asked ``Does the high-level semantic fidelity change between the two images''. They can select either ``Yes'' or ``No'' as answers. An example of our interface is shown in Figure \ref{fig:placeholder}. After selection, they are shown the next images pairs. To ensure the quality of annotations, we also insert trap questions that contain obvious semantic or no semantic changes. The data of users who consistently fail on these trap questions are discarded. The whole process leads to 723 image pairs with each annotated by a minimum of 12 independent users. For each HR-GT image pair, we average the user scores as a final indication of high-level fidelity score, ranging in $[0, 1]$, and a higher score indicates more obvious fidelity changes.



\subsection{Fidelity Evaluation of the SR Models}
\begin{figure}
    \centering
    \includegraphics[width=1\linewidth]{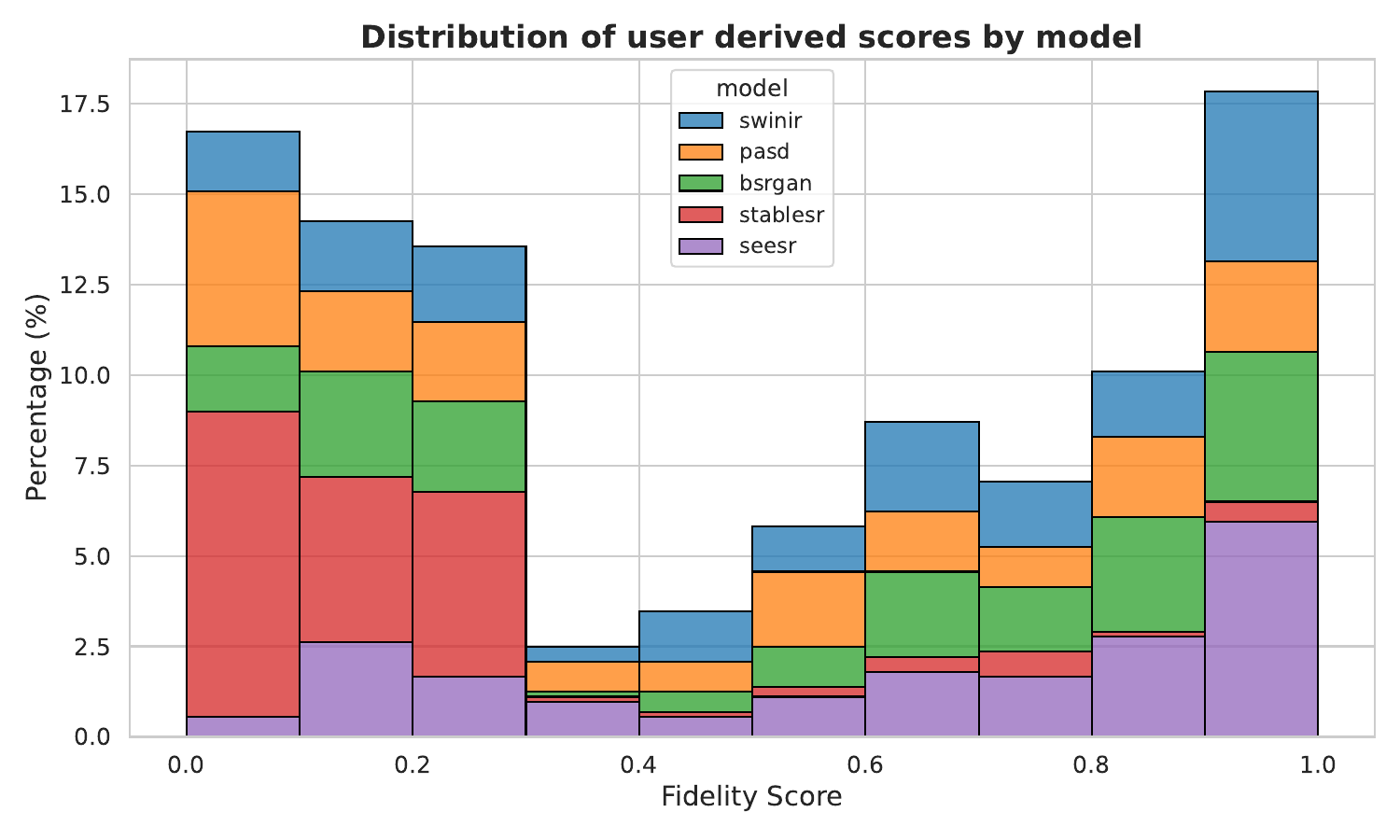}
    \caption{Statistical distribution of the fidelity scores in our  dataset. Lower means better. The fidelity scores of different SR models are labeled in different colors.}
    \label{fig:userstudyscoresdist}
\end{figure}

Figure \ref{fig:userstudyscoresdist} showcases the distribution of the fidelity scores of our constructed dataset. From the figure, we can make several observations. First, the scores follow a U-shaped distribution, indicating that different users perceive the SR model fidelity in a fairly consistent manner. This low-ambiguity feature also poses new challenges to SR models, namely that fidelity is generally either subtle or obvious to perceive; thus, it is vital to reduce high fidelity alterations when developing or optimizing future SR models. Second, among the three diffusion-based SR models, the well-performing model SeeSR \cite{wu2024seesr}, although achieving high perceptual quality in previous evaluations \cite{su2025rethinking,chen2025toward}, does not achieve a good performance in maintaining fidelity. We attribute this to their text captioning model operated on LR images, which leads to erroneous controls to the generation process once it fails to caption correct contents. As a comparison, an earlier diffusion-based SR model StableSR \cite{wang2024exploiting}, though showing relatively inferior perceptual quality, maintains most of the original information and achieves mostly low fidelity changing scores. This further raises the challenge of balancing the trade-off between the generation power and the content maintenance ability of the SR models. Last, two GAN based models BSRGAN \cite{zhang2021designing} and SwinIR \cite{Liang2021SwinIRIR} also showed high fidelity changing scores, this is due to the lack of details from the overly blurred LR input, leading to vague and indistinguishable structures of the HR output, showing both low visual quality and fidelity perceived by the users.

Finally, we split the dataset into a training and a testing set comprising 80\% and 20\% images respectively. The dataset will be further used to evaluate and develop fidelity metrics, as will be shown in the next section.


\section{Measuring High-Level Fidelity}
\begin{table}[t!]
\centering
\caption{Evaluation of different image quality metrics on the proposed high-level fidelity measurement task. SRCC and PLCC results are reported. Evaluations performed on the high-level fidelity dataset test set.}
\begin{tabular}{l c c c c c}
\hline
\textbf{Metric} & \textbf{SRCC} & \textbf{PLCC} & \textbf{Type}\\
\hline
PSNR        & 0.4505 & 0.4493 & FR  \\
SSIM        & 0.5356 & 0.5315 & FR \\
VIF         & \textbf{0.8063} & 0.7508 & FR \\
LPIPS       & 0.7681 & 0.7618 & FR \\
DISTS       & 0.7948 & \textbf{0.8040} & FR \\
A-FINE      & 0.3490 & 0.2286 & FR \\
A-FINE fidelity term & 0.7153 & 0.7541 & FR\\
A-FINE naturalness term & 0.2755 & 0.2806 & FR\\ \hline
CLIP-IQA    & 0.1367 & 0.1289 & NR \\
KONIQ++ blur & 0.2934 & 0.3229 & NR\\
KONIQ++ artifacts & 0.0529 & 0.0128 & NR \\
KONIQ++ quality score & 0.0524 & 0.0472 & NR\\
JOINT rationality score & 0.0972 & 0.0907 & NR \\
JOINT technical quality score & 0.0379 & 0.0239 & NR\\
JOINT naturalness score & 0.0975 & 0.0867 & NR \\
\hline
\end{tabular}
\label{tab:iqa_metric_results}
\end{table}In this section, we are interested in finding out metrics that can faithfully measure the high-level fidelity consistency of SR models. Therefore, we investigate: (1) Existing image quality metrics, and (2) Foundation vision models that are capable of extracting high-level information.

\subsection{Measuring Fidelity Consistency using IQA Metrics}

We evaluate a wide range of existing image quality metrics that are used in current SR model performance evaluations. The metrics cover both full-reference (FR) IQA and no-refenrece (NR) IQA categories. The selected FR-IQA metrics include pixel domain metrics---PSNR and SSIM \cite{wang2004ssim}, a wavelet domain metric---VIF \cite{vif}, deep feature domain metrics---LPIPS \cite{zhang2018lpips} and DISTS \cite{ding2020iqa}, and one recent IQA metric---A-FINE \cite{chen2025toward}---that relaxes the perfect quality assumption of GTs (including the fidelity and naturalness terms from A-FINE). NR-IQA metrics include a foundation-based model based---Clip-IQA \cite{WangCL23_clipiqa}, the blur, artifact and overall quality predictions from KonIQ++ \cite{su2021koniqplusplus}, and rationality, technical quality, and overall naturalness scores from JOINT \cite{Chen_Studyof}, which is one recent model that evaluates the rationality and naturalness of AI generated images.

We evaluate the Spearman Rank Correlation Coefficient (SRCC) and Pearson Linear Correlation Coefficient (PLCC) between IQA predictions and the annotated fidelity scores. The results are shown in Table \ref{tab:iqa_metric_results}, revealing several key findings. First, none of the NR-IQA methods align with fidelity scores, which is expected since measuring fidelity requires referencing ground truth images. Second, FR-IQA metrics including VIF \cite{vif}, LPIPS \cite{zhang2018lpips}, DISTS \cite{ding2020iqa}, and the A-FINE fidelity term \cite{chen2025toward} achieve moderately consistent predictions. This indicates that image visual quality and fidelity are not entirely independent criteria but are somewhat related in human perception. Third, the traditional metric VIF \cellcolor{} performs comparably or even better than some deep learning-based methods, which we attribute to: (1) potential overfitting risks in some deep learning IQA metrics, and (2) successful simulation of the human visual system based on information theory. Finally, while some IQA metrics achieve reasonable consistency, higher prediction accuracy can be obtained using foundation models that extract more high-level information, as will be shown in the following subsection.

\begin{figure}
    \centering
   \includegraphics[width=1\linewidth]{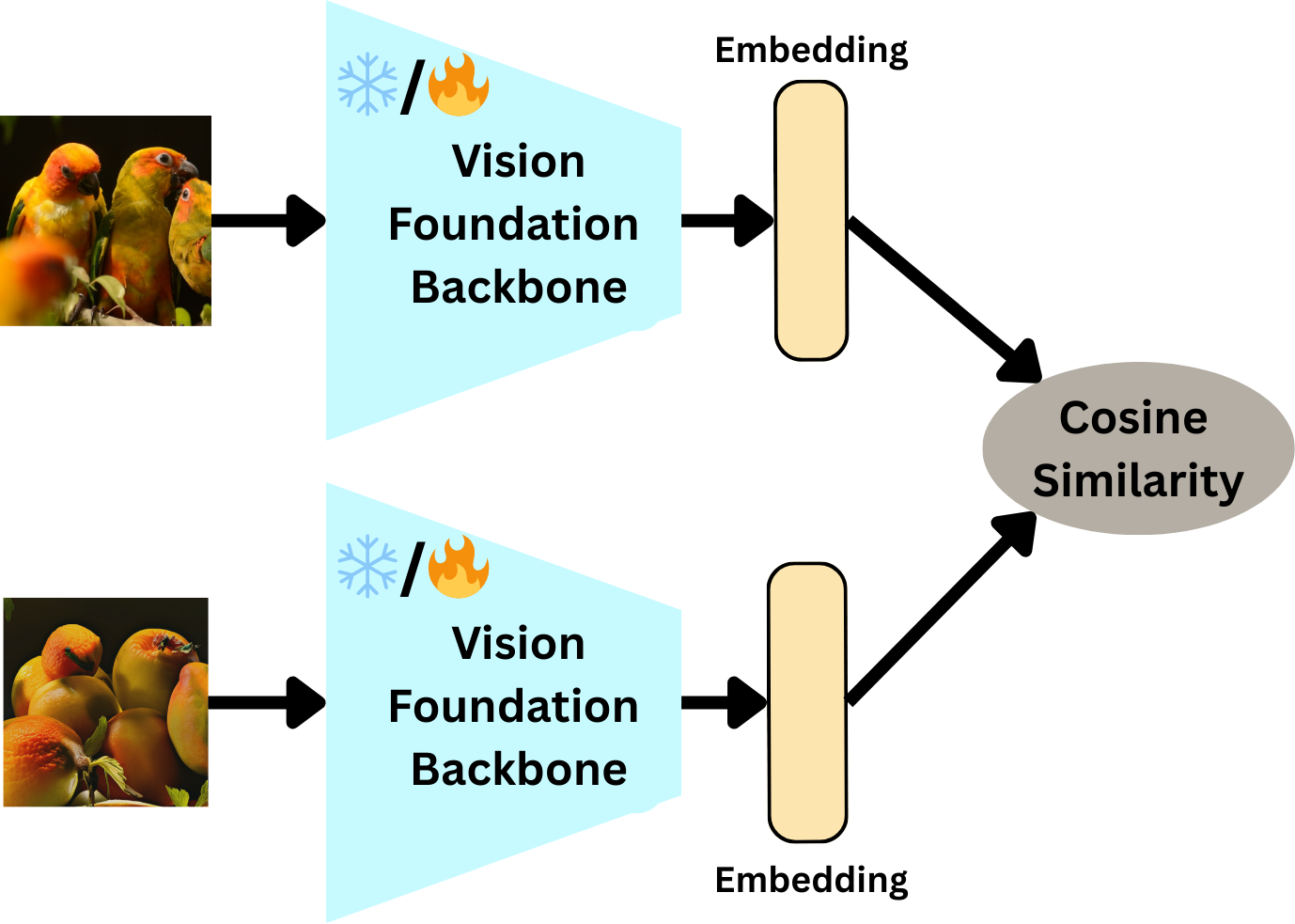}
        \caption{Method scheme. We initially pass the two images, SR and GT through the backbone and then compute the cosine similarity from the output embeddings, during training we regress this score to the dataset ground truth.}\label{fig:backbones_finetune_scheme}
\end{figure}

\begin{table}[t!]
\caption{Comparison of different foundational model embeddings in terms of Spearman Rank Correlation Coefficient (SRCC), and Pearson Rank Correlation Coefficient (PLCC), both with and without fine-tuning. Evaluations performed on the high-level fidelity dataset test set.}
\centering
\begin{tabular}{l c c c c c}
\hline
\textbf{Metric} & \textbf{SRCC} & \textbf{PLCC} \\
\hline
ViT Score   & 0.7650 & 0.8506 \\\hline
CLIP-vit-p32   & 0.8006 & 0.8539\\
BLIP & 0.7889 & 0.8544  \\
DINOv2 & 0.8008 & 0.8889\\
PE-core & 0.8045 & 0.8790  \\ \hline
(ft) CLIP-vit-p32   & 0.8384 & 0.9001  \\
(ft) BLIP & \textbf{0.8417} & 0.8955  \\
(ft) DINOv2 & 0.8315 & \textbf{0.9030}\\
(ft) PE-core & 0.8300 & 0.9017 \\
\hline
\end{tabular}
\label{tab:backbones_ft_results}
\end{table}

\subsection{A new high-level fidelity aware quality proposal}\label{sec:newmetric}
In this subsection, our aim is to derive a metric that correlates better than the above ones to evaluate high-level fidelity. Our hypothesis is that current vision foundation models already encode important semantic information; thus, we can develop a better FR metric by just comparing the embeddings of the HR image and the Ground-Truth.

As shown in ViT Score \cite{vitscore2024}, image semantic similarity can already be reflected by simply computing the cosine similarity between image features using an ImageNet-pretrained ViT, therefore, in our proposal, we also compute the cosine similarity between embeddings---in our case, the embeddings of the SR image and the Ground-truth. To further validate the effectiveness of foundation models, we test it with four different foundation models: CLIP \cite{clip2021}, BLIP \cite{li2022blip}, DINOv2 \cite{oquab2023dinov2}, and PE-core \cite{bolya2025PerceptionEncoder}. More in detail, we fine-tune all the backbones' weights on the training set of our dataset by regressing the high-level fidelity scores. We fine-tune all the models using the Adam optimizer and a learning rate of 1e-5. A schematic of our approach is shown in Figure \ref{fig:backbones_finetune_scheme}.

\begin{figure*}[t!]
    \centering
\includegraphics[width=1\linewidth]{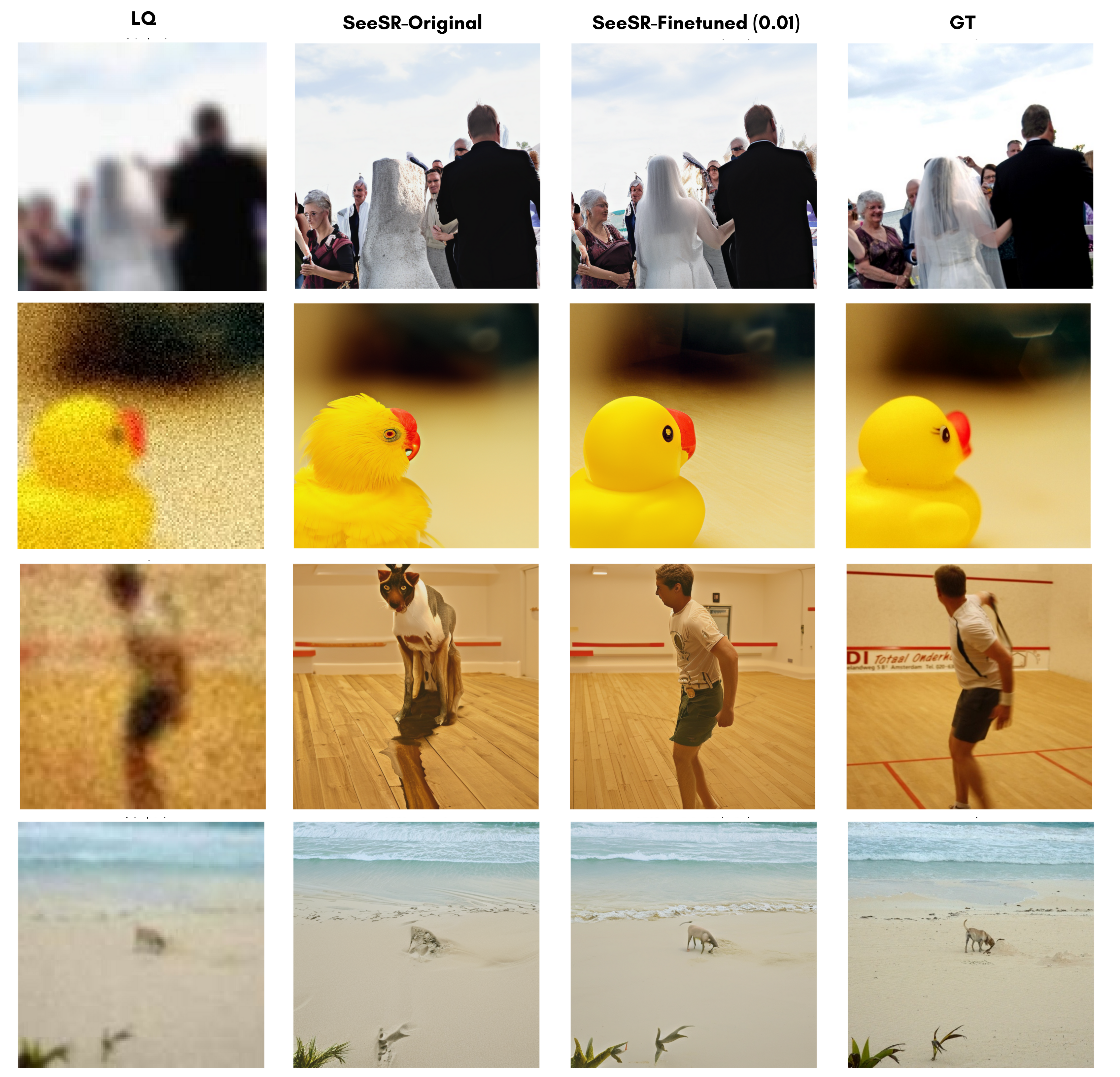}   \caption{Results for our fine-tuned SR model. From left to right, we present the original LQ image, the results of the baseline SeeSR model, our fine-tuned version with $\alpha=0.01$, and the Ground-Truth. In the first row, the baseline SeeSR model mistakenly reconstructs the bride as a rock, while our method preserves the semantic content and correctly outputs a bride. In the second row, SeeSR changes a rubber duck into a rubber parrot, whereas our approach retains the original object. In the third example, SeeSR fails to generate a recognizable person, producing a distorted output, while our model accurately reconstructs a man. Finally, in the last row, SeeSR misinterprets the dog as a rock, whereas our method faithfully reproduces the animal.}
\label{fig:seesr_improved_examples}
\end{figure*}

\begin{figure}[t!]
    \centering
\includegraphics[width=1\linewidth]{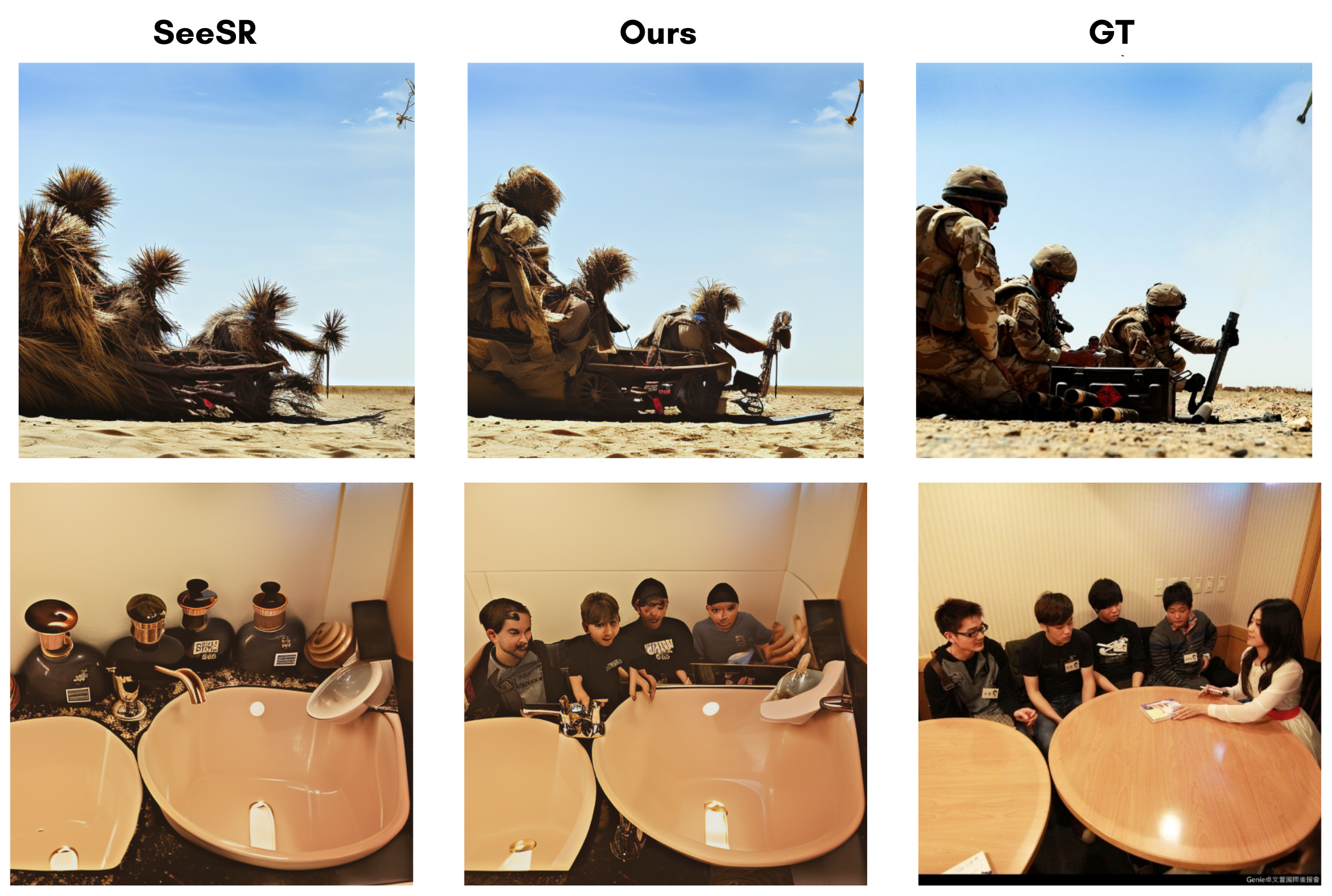}
\caption{Example of current limitations. Top: Our finetuned model successfully retrieves the two missing persons and the flag in contrast to the original model. We still encounter though some hallucinations with the second flag being a lamp. In the second example we find that even though it managed to improve the high-level fidelity of the image, some small details such the sink still present in the image.}
\vspace{-3mm}    \label{fig:limitations}
\end{figure}

Table \ref{tab:backbones_ft_results} shows the results. As before, we have computed the SRCC and PLCC. The first thing to note is that all the foundational backbones we used gain improvements over ViT Score, outlining the capabilities of these foundation models. Second, we can see that our fine-tuned models outperform all metrics in Table \ref{tab:iqa_metric_results} by up to 3\% units in SRCC and 10\% units in PLCC, proving their ability for uses as metrics to evaluate high-level fidelity consistency.

\begin{table*}[t!]
\caption{Comparison of our fine-tuned models against the original SR  across semantic and perceptual metrics. HLF stands for High-level fidelity. Best results in red and bold-faced, second best in blue and underlined. Our fine-tuned model with $\alpha=0.01$ outperforms the original SeeSR for all the metrics.}
\centering
\begin{tabular}{lcccccccc}
\hline
\textbf{Model Name} & \textbf{CLIP HLF} & \textbf{BLIP HLF} & \textbf{PE HLF} & \textbf{PSNR} &\textbf{SSIM} & \textbf{VIF} & \textbf{LPIPS} & \textbf{DISTS} \\ \hline \hline
SeeSR-Original & 0.434 & 0.478 & 0.459 & 20.746 &  0.569 & 0.058 & \cellcolor{myblue}\underline{0.399} & \cellcolor{myblue}\underline{0.249}\\



SeeSR-Finetuned (0.1)  & \cellcolor{myblue}\underline{0.416} & \cellcolor{myblue}\underline{0.459} & \cellcolor{myblue}\underline{0.450} &  \cellcolor{myblue}\underline{21.018} & \cellcolor{myblue}\underline{0.587} & 
\cellcolor{myblue}\underline{0.059} & \cellcolor{myblue}\underline{0.399} & 0.251\\

SeeSR-Finetuned (0.01)  & \cellcolor{myred}\textbf{0.412} & \cellcolor{myred}\textbf{0.456} & \cellcolor{myred}\textbf{0.436} &  \cellcolor{myred}\textbf{21.082} & \cellcolor{myred}\textbf{0.588} &  
\cellcolor{myred}\textbf{0.060} & \cellcolor{myred}\textbf{0.395} & \cellcolor{myred}\textbf{0.248}\\
\hline
\end{tabular}
\label{tab:clip_blip_perceptual}
\end{table*}

\section{Semantic fidelity aware SR models}
In the previous section, we derived a quality metric that distinguishes to a large extent whether users encounter a semantic error. Thus, the question that naturally arises is: \textit{ How to use the proposed metric to improve current Super-Resolution methods?}.

The most natural way to add the proposed metric is by incorporating it as a loss function for the minimization or fine-tuning of Super-Resolution models. In this section, we test its effect as a feedback for a recent diffusion-based model, SeeSR\cite{wu2024seesr}, which is proven to be prone to produce high-fidelity changes, as shown in Figure \ref{fig:userstudyscoresdist}.

We fine-tune the SeeSR model following:
\begin{equation}
\mathcal{L} = \mathcal{L}_{\text{ori}} + \alpha \cdot \mathcal{L}_{\text{HLF}}(I_{HR},I_{GT})
\label{eq:loss}
\end{equation}

where $I_{GT}$ is the ground-truth image, $I_{HR}$ is the result of the super-resolution model, $\mathcal{L}_{\text{ori}}$ is the original loss used in the SeeSR model, and $\mathcal{L}_{\text{HLF}}$ is our metric defined in Section \ref{sec:newmetric}---specifically the Dinov2 fine-tuned one.
We select two values of the weighting parameter, $\alpha \in \{0.1, 0.01\}$. Fine-tuning is performed on a 65k subset of the COCO 2017 training set \cite{caesar2018coco} following exactly the same pipeline as the original SeeSR. Training runs for 10 epochs, using the AdamW optimizer with a learning rate of $1 \times 10^{-5}$. We train the model for the classic $\times4$ scale.

We evaluate our two fine-tuned versions against the original SeeSR model, both on our newly proposed high-level fidelity metrics and traditional IQA metrics. Given that we use the Dinov2 fine-tuned version for training, we evaluate high-level fidelity using the three other fine-tuned versions: CLIP fine-tuned, PE-core fine-tuned, and BLIP fine-tuned. For IQA metrics, we select varying metrics, including PSNR, SSIM \cite{wang2004ssim}, VIF \cite{vif}, LPIPS \cite{zhang2018lpips}, and DISTS \cite{ding2020iqa}, covering both distortion and perception measurements. The evaluation is performed on another 1,000 images sampled from KonIQ-10k \cite{koniq10k}, following the same preprocessing procedure to construct LR images, as used in the high-level fidelity dataset. 

We report the results in Table \ref{tab:clip_blip_perceptual}. We highlight \colorbox{myred}{\textbf{best}} and \colorbox{myblue}{\underline{second-best}} values for each metric. We can see that both of our fine-tuned versions outperform the original SeeSR for the three high-level fidelity metrics, with the model with $\alpha=0.01$ obtaining better results. Moreover, our two fine-tuned versions also outperform the original SeeSR for PSNR, SSIM, and VIF, again with the model with $\alpha=0.01$ being better in this case. This model ($\alpha=0.01$) is also better than the original SeeSR for VIF and LPIPS. Thus, the results in this Table show that considering our proposed metric as an additional loss for fine-tuning current models allows for the improvement not only of the high-level fidelity error, but also helps improve the results in common image processing metrics.

Figure \ref{fig:seesr_improved_examples} shows some visual results for the original SeeSR and one of our fine-tuned models. From left to right, we show the original LQ image, the results for the original SeeSR, our fine-tuned version with $\alpha=0.01$, and the Ground-Truth. In the first row, the original SeeSR model erroneously reconstructs the bride as a rock, whereas our method successfully preserves the semantic content and outputs a bride. In the second row, SeeSR transforms a rubber duck into a rubber parrot, while our approach outputs the original object. In the third example, SeeSR fails to output a person, producing a distorted result, whereas our model correctly outputs a man. Finally, in the last row, SeeSR misrepresents the dog as a rock, while our method faithfully reconstructs the animal.



\subsection{Limitations}

Figure \ref{fig:limitations} shows some limitations of our fine-tuned model. In the first row, we can still see some hallucinated ``grass'' although the structural consistency is improved compared to the original SeeSR model. This is probably due to less concentration on the low-level textures, leading to regional fidelity errors. The second row exhibits similar limitations, where the model becomes capable of reconstructing students, however, fails to recover other details such as the sink in the image. We also point out weaknesses in reconstructing face structures, as can be found from the second example, which can also be interpreted as a high-level fidelity loss. However, we attribute this to the limited generation ability of diffusion-based model, and improving diffusion models in generating more faithful faces can be a solution.

\section{Conclusions}
Recent SR models exhibit string generative capabilities and deliver visually appealing results. However, they often compromise semantic consistency, leading to hallucinated content that is not adequately captured by existing evaluation methods.
To address this issue, we constructed the first annotated dataset with fidelity scores across multiple SR models, enabling a systematic assessment of their ability to preserve high-level image semantics. Our analysis reveals that state-of-the-art SR models vary significantly in fidelity performance, highlighting the limitations of relying solely on traditional quality metrics.
We further investigated the correlation between fidelity and diverse image quality metrics, and demonstrated that foundation models offer superior capabilities in capturing high-level fidelity. Leveraging this insight, we fine-tuned an SR model using a new fidelity-based metric, achieving improvements not only in fidelity quality metrics but also in perceptual quality ones.
Our new high-level fidelity metric provides a valuable tool for both model evaluation and optimization, and we advocate for its adoption in future SR research to ensure more reliable and semantically faithful image reconstruction.


{
    \small
    \bibliographystyle{ieeenat_fullname}
    \bibliography{main}
}

\end{document}